%% file: main_acl.tex
\pdfoutput=1

\documentclass[11pt]{article}

\usepackage{acl}

\usepackage{times}
\usepackage{latexsym}

\usepackage[T1]{fontenc}

\usepackage[utf8]{inputenc}

\usepackage{microtype}
\input{math_commands.tex}

\DeclareMathOperator*{\softmaxA}{softmax}
\usepackage{mathabx}
\usepackage{subfig}
\usepackage{float}
\usepackage{bbm}

\usepackage{times}
\usepackage{soul}
\usepackage{url}
\usepackage{hyperref}
\usepackage[utf8]{inputenc}
\usepackage{caption}
\usepackage{graphicx}
\usepackage{multirow}
\usepackage{amsmath}
\usepackage{amsthm}
\usepackage{booktabs}
\usepackage{algorithm}
\usepackage{algorithmic}
\usepackage[normalem]{ulem}
\useunder{\uline}{\ul}{}
\title{Controlling the Focus of Pretrained Language Generation Models}

\author{
Jiabao Ji\\
Shanghai Jiao Tong University\\
\texttt{jiyi0115@gmail.com}\\\And
  Yoon Kim\\
  Massachusetts Institute of Technology\\
  \texttt{yoonkim@mit.edu} \\\AND
  James Glass\\
  Massachusetts Institute of Technology\\
  \texttt{glass@mit.edu} \\ \And
  Tianxing He\\
  Massachusetts Institute of Technology\\
  \texttt{tianxing@mit.edu} \\
  }

\def\bx{{\mathbf{x}}}
\def\by{{\mathbf{y}}}
\def\bh{{\mathbf{h}}}
\def\bc{{\mathbf{c}}}

\begin{document}
\maketitle
\begin{abstract}
\vspace{-1mm}
The finetuning of pretrained transformer-based language generation models are typically conducted in an end-to-end manner, where the model learns to attend to relevant parts of the input by itself. However, there does not exist a mechanism to directly control the model's focus. This work aims to develop a control mechanism by which a user can select spans of context as ``highlights'' for the model to focus on, and generate relevant output. To achieve this goal, we augment a pretrained model with trainable ``focus vectors'' that are directly applied to the model's embeddings, while the model itself is kept fixed. These vectors, trained on automatic annotations derived from attribution methods, act as indicators for context importance. We test our approach on two core generation tasks: dialogue response generation and abstractive summarization. We also collect evaluation data where the highlight-generation pairs are annotated by humans. Our experiments show that the trained focus vectors are effective in steering the model to generate outputs that are relevant to user-selected highlights. 
\end{abstract}

\vspace{-3mm}
\input{sec_intro}
\vspace{-3mm}
\input{sec_model_formulation}

\input{sec_datasets}

\input{sec_exp}

\input{sec_related}

\vspace{-1mm}
\section{Conclusion}
\vspace{-1mm}

In this work we propose the focus vector framework as a light-weight solution to control the focus of pretrained transformer models. It has two major advantages: (1) Focus vectors act as simple transformations to the embeddings in the encoder, and the transformer model is kept fixed; (2) Attribution methods are utilized to get automatic highlight labels for training focus vectors.

We test our approach on two tasks: dialogue response generation, and abstractive summarization. For evaluation, we collect data where the highlight-generation pairs are annotated by humans. Experiments show that the trained focus vectors are effective in steering the model to generate output text that is relevant to the specified highlights.

\section{Acknowledgements}
We sincerely thank Evan Hernandez, Ekin Akyürek, Joe O'Connor, Hongyin Luo, and Jacob Andreas for helpful discussions. This research was sponsored by the United States Air Force Research Laboratory and the United States Air Force Artificial Intelligence Accelerator and was accomplished under Cooperative Agreement Number FA8750-19-2-1000. The views and conclusions contained in this document are those of the authors and should not be interpreted as representing the official policies, either expressed or implied, of the United States Air Force or the U.S. Government. The U.S. Government is authorized to reproduce and distribute reprints for Government purposes notwithstanding any copyright notation herein.

\bibliography{emnlp2021}
\bibliographystyle{acl_natbib}

\clearpage
\appendix

\input{appendix}

\end{document}

%% file: math_commands.tex
\usepackage{amsmath,amsfonts,bm}

\def\eqref#1{equation~\ref{#1}}

\def\1{\bm{1}}

\DeclareMathAlphabet{\mathsfit}{\encodingdefault}{\sfdefault}{m}{sl}
\SetMathAlphabet{\mathsfit}{bold}{\encodingdefault}{\sfdefault}{bx}{n}



%% file: sec_intro.tex
\section{Introduction}
\label{sec:intro}

Transformer-based models pretrained on large-scale text data have become the dominant paradigm for natural language generation (NLG) tasks \citep{roller2020recipes,lewis2019bart,2020t5}. The attention module \citep{bahdanau2016neural,tfattention17Vaswani}, which aggregates information  via a weighted average over word-level embeddings, plays a vital role in these models.
The attention mechanism serves two major purposes: (1) It captures linguistic phenomena in the input \citep{clark-etal-2019-bert,kovaleva-etal-2019-revealing,kobayashi-etal-2020-attention}; (2) It helps the model focus on relevant portions of the input (e.g., alignment in machine translation \citep{bahdanau2016neural} and abstractive summarization \citep{rush-etal-2015-neural}).


\begin{figure}[t]
    \centering
    \includegraphics[width=0.85\linewidth]{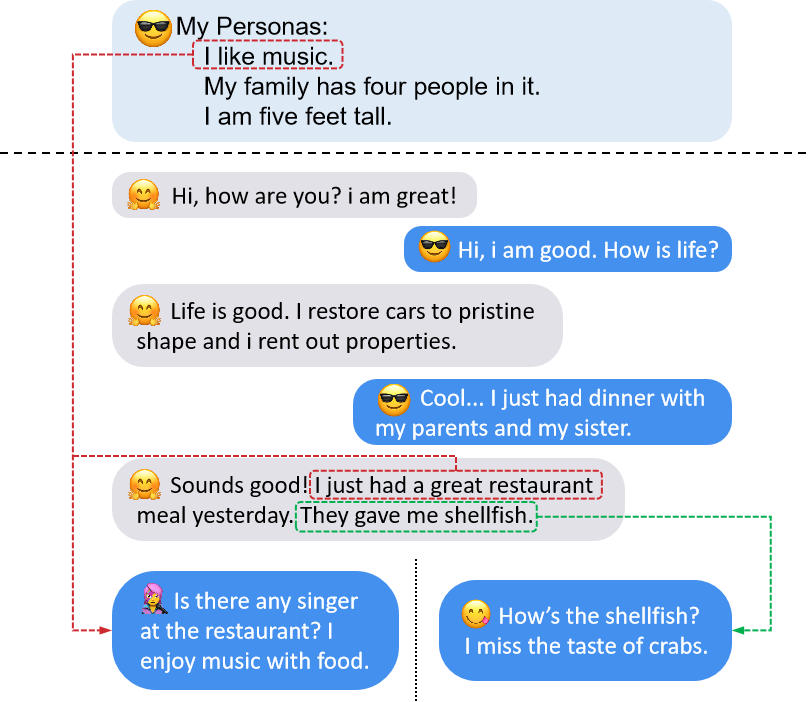}
    \vspace{-1mm}
    \caption{Illustration of our motivation: different highlights in the input (including persona) lead to different generations. This example is from our collected dialogue data for evaluation (Section \ref{sec_datasets}).}
    \label{fig:mo_diffatt}
    \vspace{-5mm}
\end{figure}

The attention module is particularly useful as it does not require any explicit supervision: the model learns to attend to relevant parts of the input \textbf{by itself} through end-to-end training. However, this property makes it difficult to explicitly control the model's focus. \textbf{If the model happens to put focus on some span of context that the user thinks is not so important, we currently do not have a mechanism to correct it.} This is especially sub-optimal in some NLG applications involving a relatively long input such as dialogue or summarization: focusing on different spans of the input could result in completely different generations (illustrated in Figure \ref{fig:mo_diffatt}). It would be attractive to give the user an option to control the model's focus.

In this work, we aim to develop a mechanism to steer the model to generate output relevant to some user-specified input spans (which we term as highlights).\footnote{To avoid confusion, our goal is \textit{not} about controlling the attention modules inside the model, instead, we care about the actual generation.} This goal, however, brings about significant challenges. For one, many popular NLG datasets are collected in an end-to-end manner, i.e., without annotations of which spans of input are most relevant to the reference target. It would also be ideal for the proposed approach to be compatible with existing pretrained transformer models, as re-training such models is often costly.




In this work, we propose an \textit{focus vector} framework to address the challenges outlined above. Our contributions are as follows:
\begin{itemize}
    \item To control the model's focus, we augment the pretrained model with trainable focus vectors which are directly applied to the encoder embeddings. The model itself is kept fixed, and no further changes to the model architecture is needed.
    \item The training of focus vectors does not require additional annotations. We utilize attribution methods to derive automatic highlight annotations from existing end-to-end training data.
    \item For principled evaluation and future work in this direction, we collect and release human evaluation data where the highlight-generation pairs are annotated by humans.
    \item We test our approach on two core NLG tasks: dialogue response generation and abstractive summarization. Experiments show that the trained focus vectors are effective in steering the model to generate a relevant output given the selected highlights.
\end{itemize}

%% file: sec_model_formulation.tex
\section{Model Formulation}
\label{sec:model}
\vspace{-1mm}
We assume the target model is a standard pretrained transformer encoder-decoder model \citep{tfattention17Vaswani} that has already been finetuned on end-to-end task-specific data (e.g., dialogue or summarization) with the standard negative log-likelihood (NLL) loss. Our goal is to establish a control mechanism whereby the user can highlight several spans of the input, and the model is supposed to generate outputs relevant to the highlighted text. Crucially, this mechanism should not change the base model, in order to allow the user to default back to the original model if desired.

We begin by establishing notation. We denote the end-to-end training data by $\{\bx,\by\}$, where $\bx=\{x_1,...,x_n\}$ refers to the input token sequence, and $\by$ refers to the corresponding reference target token sequence. During evaluation, some spans of the input $\bx$ will be highlighted, and we use a binary indicator $c_i$ to indicate whether the $i^{th}$ input token is to be highlighted during generation. In this work we only consider a set of complete sentences as a valid highlight span. This design choice is mainly for convenience during our human-annotated evaluation data collection, and our framework can readily be generalized to phrase-level highlights. 

Suppose the encoder model is composed of $L$ transformer layers. We denote the $d$-dimensional output embedding of the $i^{th}$ position on the $l^{th}$ encoder layer by $\bh^l_i$. We use  $\{\bh^0_i\}$ to denote the input embeddings. Each decoder layer performs multi-head cross-attention on the outputs of the encoder, where the attention weight computation for the $h^{th}$ head on the $l^{th}$ decoder layer is formulated as below:
\begin{equation}
\label{eq:cross_attention}
    \alpha^{h,l}_{i,j}=\softmaxA_{i \in \{1...n\}} \left( \frac{k(\bh^L_i) \cdot \mathbf{q}^{h,l}_j}{\sqrt{d}} \right).
\end{equation}
Here $k(\cdot)$ is a linear transform, and $\alpha_{i,j}$ is the attention weight assigned to encoder output $\bh^L_i$, for the $j^{th}$ position decoder query vector $\mathbf{q}_j$. We use $P_\text{M}(\by|\bx)$ to denote the probability assigned to $\by$ given input $\bx$ by the original target model. For more details of the transformer encoder-decoder architecture, we refer readers to \citet{tfattention17Vaswani}.

Our proposed framework involves two stages. We first obtain automatic highlight annotations using attribution methods. Then, these annotations are used to train the focus vectors. In the next section, we review the attribution methods. 

\vspace{-1mm}
\subsection{Attribution Methods}
\vspace{-1mm}

Many popular NLG datasets are collected end-to-end, i.e., without annotations of which spans of input are most relevant to the reference target. To obtain these annotations for focus vector training, we make use of existing attribution methods.

Attribution methods  \citep{JMLR:v11:baehrens10a,Simonyan14deepinside,pmlr-v70-shrikumar17a,NEURIPS2018_294a8ed2,pmlr-v70-sundararajan17a}, also known as \textit{saliency maps}, attribute the prediction of a (potentially black-box) model to its input features. It thus fits our need to extract relevant spans in the input given the reference target. Most saliency methods are originally designed for image classification, where an importance score is assigned for each dimension of the input feature. Therefore, slight modifications (e.g., dot-product with the word embeddings) are needed to apply them to language data  \citep{ding-koehn-2021-evaluating,DBLP:journals/corr/DenilDF14}. 

We implement and compare several popular attribution methods, which compute the attribution score for a given sentence $S$ (denoting the set of token indexes in the sentence) in the input $\bx$ for the  target $\by$ and model $P_\text{M}$. 
\vspace{-2mm}
\paragraph{Leave-one-out (LOO)} We replace the tokens in $S$ by the \texttt{<pad>} token, and compute the difference in NLL: 
\begin{equation}
A(S)=\log P_\text{M}(\by|\bx)-\log P_\text{M}(\by|\bx_{S\text{-padded}}).
\end{equation}
LOO is also referred to as an \textit{occlusion-based method} \citep{zeiler2014visualizing,DBLP:journals/corr/LiMJ16a} in the literature.
\vspace{-2mm}
\paragraph{Attention-weight} We sum up the attention weights assigned to tokens in $S$ for all attention heads across all decoder layers:
\begin{equation}
    A(S) = \sum_{i \in S}\sum_{j,h,l}\alpha^{h,l}_{i,j}. \vspace{-1mm}
\end{equation}

\vspace{-2mm}
\paragraph{Grad-norm} We sum the norm of gradient for the input word embeddings in $S$: 
\begin{equation}
A(S)=\sum_{i \in S}||\nabla_{\bh^0_i}\log P_\text{M}(\by|\bx)||_2. \vspace{-1mm}
\end{equation}

\vspace{-2mm}
\paragraph{Grad-input-product} Instead of taking vector norm, we compute the dot-product between the input embedding and its gradient:
\begin{equation}
A(S)=\sum_{i \in S} \left(\nabla_{\bh^0_i}\log P_\text{M}(\by|\bx)\right) \cdot \bh^0_i.\vspace{-1mm}
\end{equation}
While more sophisticated attribution method have been proposed in the literature \citep{lei-etal-2016-rationalizing,pmlr-v70-sundararajan17a,bastings-etal-2019-interpretable}, we mainly experiment with the methods listed above due to their simplicity and popularity.  Attribution methods have been used for interpreting black-box models---\textbf{applying them to derive labels that can further be used to control the focus of a model has to our knowledge not been explored before}. 

Which attribution method best reflects the model's inner working is still an active research area \citep{ding-koehn-2021-evaluating, NEURIPS2018_294a8ed2}. The present work is primarily concerned with how well the attribution scores align with human-annotated highlights. In our experiments, we find that leave-one-out (LOO) has the best correlation on the human-annotated development set (Table \ref{tab:attribute_res}, details given in Section \ref{sec_datasets}). We therefore adopt LOO to derive the automatic highlight annotations. 

More specifically, for the input-output pairs in the training set, we sort the LOO attribution scores of the sentences in the input from large to small, and mark the tokens in the first few sentences (the exact number varies by task) as highlights. We denote the highlight labels obtained from this automatic procedure by a binary indicator variables $\mathbf{c}^\text{attr}=\{c_1^{\text{attr}}, \dots, c_n^{\text{attr}}\}$, which will be used to train the focus vectors.

\begin{figure}
    \centering
    \includegraphics[width=0.9\linewidth]{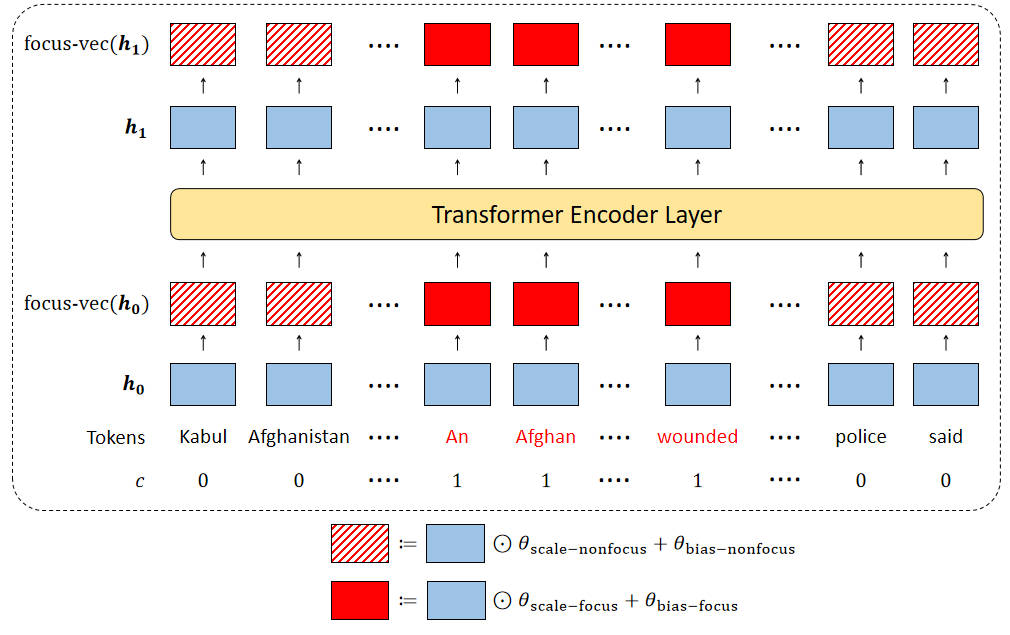}
    \vspace{-0.2cm}
    \caption{Illustration of our proposed focus vectors applied to a one-layer transformer encoder. The parameters of the transformer model are kept fixed. The highlighted spans are filled by red.}
    \label{fig:attvec_illustrate}
    \vspace{-0.4cm}
\end{figure}

\vspace{-1mm}
\subsection{Focus Vectors}
\vspace{-1mm}

To control the model's focus, we introduce a set of $d$-dimensional vectors $\theta$, named \textit{focus vectors}. They are designed to act as \textit{indicators} for the model, designating which parts of the input to focus on. We now assume the training set contains $\{\bx, \mathbf{c}^\text{attr}, \by\}$ triples, where $\mathbf{c}^\text{attr}$ is obtained from the attribution method from the previous section. Focus vectors modify the forward pass of the encoder model by applying a simple transformation $f$ on the output embeddings of each layer (including the input layer):
\begin{equation}
\begin{split}
\label{eq:att_vec}
\small
    f(\bh^l_i) = \begin{cases} 
        \bh^l_i \odot \theta^{l}_\text{scale-focus} + \theta^{l}_\text{bias-focus}, & \text{if}~c^\text{attr}_i = 1 \\
        \bh^l_i \odot \theta^{l}_\text{scale-nonfocus} + \theta^{l}_\text{bias-nonfocus}, & \text{if}~c^\text{attr}_i = 0
        \end{cases}. \\
\end{split}
\end{equation}
We provide an illustration in Figure \ref{fig:attvec_illustrate}. The total number of parameters introduced by the focus vectors is therefore $4 \times (L + 1) \times d$, which is negligible in comparison to the large number of parameters of the fixed transformer model.
We note that as the focus vectors operate directly on the encoder embeddings, it does not require an explicit attention module to exist in the model and is therefore applicable to non-attentional architectures such as LSTMs \citep{DBLP:journals/corr/HuangXY15}.

We train the focus vectors using the standard NLL loss with stochastic gradient descent (SGD):
\begin{equation}
\label{eq:attvec_loss}
    \mathcal{L(\bx,\by,\bc^\text{attr};\theta)} = - \log P_\text{focus}(\by|\bx,\bc^\text{attr}),
\end{equation}
where $P_\text{focus}(\cdot|\bx,\bc^\text{attr})$ denotes the distribution over the output after the focus vectors are applied. We re-iterate that during training of the focus vectors, the transformer model is kept fixed. This allows the user to default back to the pretrained model (i.e., without applying the focus vectors), if the user prefers not to specify any highlights. 

Readers may wonder what is the difference between our approach and standard end-to-end training, as  both cases use the same $\bx,\by$ pairs. This is related to our key assumption that \textit{different focus of the input lead to different generations}, and the fact that $\bc^\text{attr}$ is the relevant span for $\by$ in the ideal case. Therefore, the focus vectors have the opportunity to give information about which span is more relevant to $\by$, before the model observes $\by$ on the decoder side. To reduce the loss $-\log P_{\text{focus}}(\by|\bx, \bc^\text{attr})$, the focus vectors need to steer the model's focus towards the spans marked by $\bc^\text{attr}$.  

At test time, the user will highlight several sentences in the input which we denote by $\bc^\text{user}$. We apply the trained focus vector according to Equation \ref{eq:att_vec}, and decode the output from $P_\text{focus}(\cdot|\bx,\bc^\text{user})$.

%% file: sec_datasets.tex
\section{Evaluation Data Collection}
\label{sec_datasets}
\vspace{-1mm}


We test our method on two NLG tasks: dialogue response generation and abstractive summarization. For the dialogue task, we adopt the \textit{PersonaChat} dataset \citep{zhang2018personalizing}. It is an open domain multi-turn chit-chat dataset, where two participants are required to get to know each other by chatting naturally. Each of them is given a \textit{persona}: several pieces of personal information such as \textit{``I major in  Computer Science''}, serving as background information. The participants are required to reflect their assigned persona in the conversation. For  summarization, we adopt the \textit{CNN/Dailymail} dataset \citep{hermann2015teaching,nallapati2016abstractive}, which is a standard dataset for end-to-end abstractive summarization. To save space, we defer details and statistics of the datasets to Appendix \ref{sec:app_datasets}.



\begin{table}[]
\small
\centering
\begin{tabular}{@{}c|clcl@{}}
\toprule
\textbf{Attribution Method} &
  \multicolumn{2}{c}{\begin{tabular}[c]{@{}c@{}}\textbf{PersonaChat}\\ \textbf{P@1(\%)}\end{tabular}} &
  \multicolumn{2}{c}{\begin{tabular}[c]{@{}c@{}}\textbf{CNN/Dailymail}\\ \textbf{P@1(\%)}\end{tabular}} \\ \midrule
\multicolumn{1}{c|}{attention-weight} & \multicolumn{2}{c}{29.18} & \multicolumn{2}{c}{40.31} \\
\multicolumn{1}{c|}{grad-norm}       & \multicolumn{2}{c}{54.00} & \multicolumn{2}{c}{43.87} \\
\multicolumn{1}{c|}{grad-input-product}      & \multicolumn{2}{c}{44.05} & \multicolumn{2}{c}{32.60} \\
\multicolumn{1}{c|}{leave-one-out} & \multicolumn{2}{c}{\textbf{62.31}} & \multicolumn{2}{c}{\textbf{64.43}} \\ \bottomrule
\end{tabular}
\vspace{-1mm}
\caption{Top-1 precision (\%) of different attribution methods on the human-labeled development set.}
\label{tab:attribute_res}
\vspace{-2mm}
\end{table}

Both PersonaChat and CNN/Dailymail are created end-to-end and do not contain annotated highlight spans. \textbf{For principled evaluation, we utilize the Amazon Mechanical Turk (AMT) platform to collect evaluation sets where the highlight-generation pairs are annotated by humans.}

For PersonaChat, each turker\footnote{We recruit turkers located in the U.S.} is shown a dialogue history and the corresponding persona of the speaker. The dialogue history is randomly selected from the original test set of PersonaChat. Then the turker is required to choose 1-3 sentences as highlights (for example, one sentence in persona, and one sentence in dialogue history), and write down a dialogue response that not only continues the current dialogue, but also is relevant to the chosen highlights. Finally, we ask the turker to repeat the above process, but select a different set of highlights and provide another response. After a few preliminary trials and modifications to our instructions / rewards, we find that turkers comply nicely with our instructions and provide high-quality highlight-response pairs.

For CNN/Dailymail however, we first found that turkers had difficulty writing a high-quality summary for a given news article, with many turkers giving random responses even after we increased the reward. This is perhaps unsurprising given that writing a good summary is challenging and the reference summaries are written by experts. After a few disappointing iterations, we turn to a compromise: we directly provide the turkers with the reference summary, and only ask them to select 2-5 relevant sentences in the article. This simplifies the task, and we are able to collect high-quality labels. This compromise is not ideal, as it reverses the order of highlighting and generation. However, we find that in most cases, the reference summaries in CNN/Dailymail are well covered by several ``key'' sentences in the article, which are highlighted by the turkers. Therefore, we believe this compromise does not hurt the soundness of our evaluation.

In order to ensure high data quality for both dialogue and summarization, we design a qualification test that turkers need to pass before conducting the actual tasks. Several automatic checks and a minimal time limit are added in the scripts to prevent trivial answers. We also manually monitor the incoming submissions, and ban misbehaving turkers and discard their submissions. More details about our AMT setup are provided in Appendix \ref{sec:app_eval}.

Our final collected datasets include 3,902 highlight-generation pairs for PersonaChat, and 4,159 pairs for CNN/Dailymail. They are randomly split 50/50 into dev/test sets. \textbf{We include a number of samples of our collected data in the supplementary materials.} Our code and the collected dataset will be released in \url{https://github.com/Question406/LearningToFocus}.\footnote{We will make it ready approximately in early April 2022.} We hope that this evaluation data could facilitate future research in this direction.

\vspace{-1mm}
\paragraph{Comparison of Attribution Methods} We use the collected highlight-generation pairs in the dev set to compare which attribution method aligns best with human-annotated highlights. In particular, we compute the top-one precision of the sentence ranked highest by the attribution method. The result is shown in Table \ref{tab:attribute_res}. We find that for both PersonaChat and CNN/Dailymail, LOO has the best alignment. We therefore use LOO to obtain automatic annotations for focus vector training. Interestingly, we observe low alignment between attention weight-derived attribution scores and human judgment, which indicates that controlling model generations via intervening on the attention distributions may not optimal. Finally, we note  that this result does not mean LOO is the ``best'' attribution method, as attribution method is supposed to reflect the model's inner working, instead of a human's.

%% file: sec_exp.tex
\section{Experiments}

\vspace{-1mm}
\subsection{Experiment Setting and Baselines}
\vspace{-1mm}

We use Blenderbot \citep{roller2020recipes} as the base model for PersonaChat and BART \citep{lewis2019bart} for CNN/Dailymail, both of which are standard encoder-decoder transformer models. Our code is based on the \textit{transformers} library \citep{wolf-etal-2020-transformers}. We load the pretrained weights from \texttt{facebook/blenderbot-400M-distill} and \texttt{facebook/bart-base}. Blenderbot has 2 encoder layers and 12 decoder layers, while BART  has 6 encoder layers and 6 decoder layers. To help  Blenderbot cope with long dialogue context in PersonaChat, we extend its maximum position embedding index from 128 to 256.
We use beam-search for decoding, where we  follow the recommended configuration \citep{roller2020recipes, lewis2019bart} and use a beam size of 10 for Blenderbot and a beam size of 4 for BART.

For both tasks, we first finetune the base model on the original training set in the standard end-to-end manner. The model is then fixed and used to obtain automatic labels $\bc^\text{attr}$ with the LOO attribution method on the same training set. For each training sample, we select the top-$k$ sentences in the input ranked by LOO. Since we do not know the best value for $k$, we set it to be a random number from 1 to 3 for PersonaChat, and from 2 to 5 for CNN/Dailymail.

While the highlight labels in the training set used to train focus vectors are derived automatically, we use the human-labeled dev set for hyper-parameter tuning. This is to facilitate fair comparison with other baseline approaches which also utilize the human-labeled dev set. In our ablation study, we will show that this dependence on human-labeled dev set is not crucial for our approach to achieve strong performance.
We perform a grid search over learning rate with $\{1, 3, 5\} \times \{1e^{-4}, 1e^{-3}, 1e^{-2}, 1e^{-1}\}$. The Adam optimizer \citep{kingma2014adam} is used with $\beta_1 = 0.9, \beta_2=0.999$, and a L2 decay weight of $0.01$. For both tasks, we set the mini-batch size to be 16.

\begin{table*}[t]
\small
\centering
\begin{tabular}{@{}c|ccc|ccc@{}}
\toprule
\multirow{2}{*}{\textbf{Model}}        & \multicolumn{3}{c|}{\textbf{PersonaChat}}                       & \multicolumn{3}{c}{\textbf{CNN/Dailymail}} \\ 
                             & \textbf{PPL}   & \textbf{ROUGE-1/2/L}      & \textbf{BERTScore}                  & \textbf{PPL}   & \textbf{ROUGE-1/2/L}  & \textbf{BERTScore}  \\ \midrule
\multicolumn{1}{c|}{vanilla (w.o. highlight)}    & 28.73 & 17.02/2.73/14.52 & \multicolumn{1}{c|}{85.41} & 4.51 & 43.48/21.01/30.98 & 89.23 \\
\multicolumn{1}{c|}{padding} & 38.93 & 16.69/2.80/13.72 & \multicolumn{1}{c|}{84.42} & 19.62      &     39.31/18.44/28.67 &     88.34       \\
\multicolumn{1}{c|}{keyword-control}    & 23.64 & 17.31/3.02/14.81 & \multicolumn{1}{c|}{85.58} & 4.56 & 43.81/21.08/31.15 & 89.26 \\
\multicolumn{1}{c|}{attention-offset} & 23.79 & \textbf{21.10}/3.77/17.54 & \multicolumn{1}{c|}{86.04} & 4.49 & 43.96/20.64/31.26 & 89.28 \\
\multicolumn{1}{c|}{\textbf{focus-vector}}  & \textbf{22.51} & 20.81/\textbf{3.98}/\textbf{17.58} & \multicolumn{1}{c|}{\textbf{86.13}} & \textbf{4.48} & \textbf{45.92}/\textbf{23.03}/\textbf{32.98} & \textbf{89.78} \\ \bottomrule
\end{tabular}
\vspace{-1mm}
\caption{Main evaluation results on the PersonaChat and CNN/Dailymail datasets with annotated highlights. The proposed focus vector approach shows strong performance across different metrics.}
\label{tab:main_res}
\vspace{-2mm}
\end{table*}

We compare the proposed focus-vector approach with several baselines:
\vspace{-1mm}
\paragraph{Vanilla:} The vanilla model, without any modification in both the model and the input.  
\vspace{-1mm}
\paragraph{Padding:} One trivial way to control the model's focus is to replace all input by the \texttt{<pad>} token, except the spans highlighted by the user. However, we find that this direct padding during evaluation results in drastically worse perplexity. To alleviate this problem, we randomly pad a portion of sentences in the input during the standard end-to-end finetuning, to make the model aware that only partial input would be provided.
\vspace{-1mm}
\paragraph{Keyword-control:} Keyword-based prompts \citep{fan2017controllable, he2020ctrlsum} has been a popular approach for controllable text generation. We adapt this idea to our focus-control setting. During model finetuning, we preprend key-phrases extracted from the reference target sequence to the original input. We utilize Yake \cite{campos2020yake}, which is an unsupervised keyword extraction method. During evaluation, we extract and preprend key-phrases extracted from the highlighted sentences. 


\vspace{-1mm}
\paragraph{Attention-offset:} As a direct way to control the model's attention, we add a positive scalar offset $s^\text{offset}$ to the cross-attention heads before the softmax operation (Equation \ref{eq:cross_attention}), for the highlighted spans. A similar technique has been used in \citet{dong-etal-2021-fly} to \textit{modulate} the attention distribution to tackle neural text degeneration problems \citep{curious19ari}. This approach modifies the attention weights via:
\begin{equation}
\small
    \alpha'_{i,j}=\softmaxA_{i \in \{1...n\}} \left( \frac{k(\bh^L_i) \cdot \mathbf{q}_j}{\sqrt{d}} + s^\text{offset} \cdot \mathbbm{1}_{[c_i=1]} \right),
\end{equation}
where $s^\text{offset}$ is a hyper-parameter, and is applied to all cross-attention heads in the decoder. We tune $s^\text{offset}$ on the human-annotated development set in a fine-grained manner. More details are given in Appendix \ref{sec:app_implement}.

Whether the attention distribution faithfully explains a model's predictions is the subject of much debate \citep{sarthak19attnotexp,wiegreffe-pinter-2019-attention,bastings-filippova-2020-elephant}. Therefore this direct modification of the attention head may not be the optimal solution for focus control. Our proposed focus-vector framework, on the other hand, utilizes attribution methods, and directly operates on the encoder embeddings.

\subsection{Results and Analysis}
\vspace{-1mm}

During evaluation, human-annotated highlights are fed to the model.
In addition to perplexity, we evaluate the generations from different approaches using two popular NLG metrics: ROUGE \citep{lin2004rouge}, and BERTScore \citep{zhang2019bertscore}.

We show the main results in Table \ref{tab:main_res}. As expected, the padding baseline has poor performance, as a large portion of input is masked out. Comparing to various baselines, focus-vector obtains significantly improved ROUGE and BERTScore on both tasks. This validates the motivation of this work: focus-vec is effective in steering the model's focus, which leads towards the desired generation. For CNN/Dailymail, the perplexity of focus-vector is close to the vanilla model even though there is a large difference in ROUGE. We believe this is due to the constrained nature of the summarization task and how perplexity is computed: once the model observes the first few tokens, it is easy to figure out what the current highlight is. The other two metrics, on the other hand, are based on the actual generation, and therefore does not have this issue.

\begin{table*}[]
\footnotesize
\centering
\begin{tabular}{p{0.05\linewidth}p{0.40\linewidth}|p{0.40\linewidth}}
\toprule
\multicolumn{1}{r|}{persona:} &
  \multicolumn{1}{c|}{\begin{tabular}[c]{@{}c@{}}\textbf{I have a dog named pedro.} I like to eat muffins.\\  I am five feet tall.  I work in the healthcare industry.\end{tabular}} &
  \multicolumn{1}{c}{\begin{tabular}[c]{@{}c@{}}I have a dog named pedro. I like to eat muffins.\\  I am five feet tall.  \textbf{I work in the healthcare industry.}\end{tabular}} \\ \midrule
  \multicolumn{1}{r|}{context:} & Hello, how are you today? 
 & Hello, how are you today? \\ \midrule
  \multicolumn{1}{r|}{reference:} &
  Scared, i'm looking for \textcolor{red}{my dog}, his name is pedro. have you seen him? &
   Really tired, it's so busy in \textcolor{red}{healthcare industry}. I wish I can have a long vacation.  \\
  \midrule
\multicolumn{1}{r|}{vanilla:} &
  Hi! I am doing well, how about yourself? What do you do for a living? &
  Hi! I am doing well, how about yourself? What do you do for a living? \\
\multicolumn{1}{r|}{keyword:} &
  I am doing well, how about yourself? What do you like to do for fun? &
   I am doing well. how about you? What do you do for a living? I am \textcolor{red}{in healthcare}. \\
\multicolumn{1}{r|}{att-offset:} &
  I am doing well, and you? \textcolor{red}{I am petro,  my dog}, and you? &
    I am doing well, and you? I am a \textcolor{red}{healthcare worker in the health industry}. \\
\multicolumn{1}{r|}{focus-vec:} &
  I am good. Just got back from \textcolor{red}{walking my dog}. Do you have any pets? &
  I am doing well. How about you? What do you do for a living? I am \textcolor{red}{in healthcare}. \\ \bottomrule
\end{tabular}
\vspace{-1mm}
\caption{Generation samples from different attention control approaches for PersonaChat. The inputs on both sides are identical, except a different piece of persona is highlighted (marked in bold), which leads to different generations for approaches with control. Another example is provided in Table \ref{tab:persona_sample2} (Appendix \ref{ref:app_aux_sample}).}
\vspace{-1mm}
\label{tab:persona_sample}
\end{table*}

The performance of keyword-control, although better than the vanilla model, is inferior to attention-offset and focus-vector. We surmise this is due to the following two weakness: First, key-phrases can not fully represent the highlighted span. Second, there is a discrepancy of where the key-phrases are extracted between training and evaluation.

The performance gap (in ROUGE/BERTScore) between focus-vector and attention-offset is larger on the CNN/Dailymail dataset. We believe this is because the BART model has a deeper encoder than the Blenderbot model. As the encoder grows deeper, the embeddings become more ``contextualized'' and its \textit{identifiability} \citep{Brunner2020On} degrades. And since the decoder attends to the last layer of the encoder, this direct manipulation of attention weights could be ineffective with deep encoders.

Table \ref{tab:persona_sample} shows generation samples from different focus-control approaches for PersonaChat. Spans of the generation that are relevant to the highlighted persona are marked in red. Comparing to the generation from the vanilla model, the generations from both attention-offset and focus-vector are highly relevant to the respective highlighted persona. One generation from att-offset is a little erratic (\textit{``I am petro, my dog''}), which may be due to the inflexibility of att-offset.

We defer the generation examples for CNN/Dailymail to Table \ref{tab:dailymail_sample} and Table \ref{tab:dailymail_sample2} (Appendix \ref{ref:app_aux_sample}) due to space constraints. We observe that the generation from focus-vector is more focused on the highlighted inputs. On the other hand, attention-offset's generation still remains similar to the vanilla model. 



\begin{figure*}[t]
    \centering
    \includegraphics[width=0.95\linewidth]{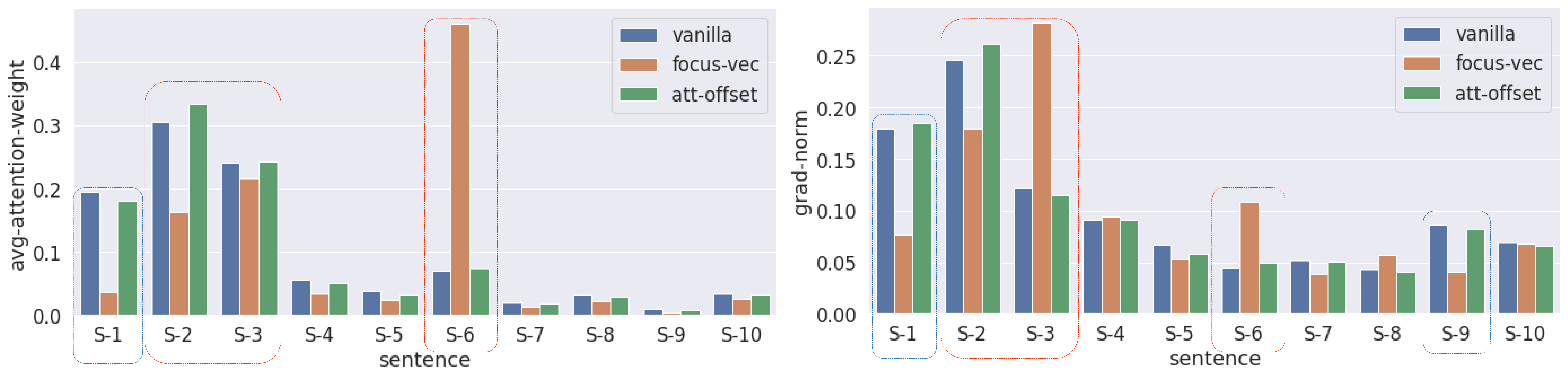}
    \vspace{-0.25cm}
    \caption{Attribution scores for each sentence in the input, with different focus-control approach applied to BART. The highlighted sentences are marked by red rectangles. The corresponding example is in Table \ref{tab:dailymail_sample} (Appendix \ref{ref:app_aux_sample}).}
    \vspace{-0.5cm}
    \label{fig:attr_analysis}
\end{figure*}

In Figure \ref{fig:attr_analysis}, we study how the outputs of attribution methods (attention-weight and grad-norm) change with different approaches (vanilla, focus-vector and attention-offset) for the CNN/Dailymail example (Table \ref{tab:dailymail_sample}). Note that in this analysis, for the attribution methods we set the target $\by$ to be the decoded output from the respective modeling, instead of the reference summary. The highlighted sentences are marked by the red rectangles.

We observe that for both attention-weight and grad-norm, the application of focus vector makes the highlighted sentences obtain the highest attribution scores, and the scores differ significantly from the vanilla model. In some of the non-highlighted sentences (marked by the blue rectangles),  attention-offset is not strong enough to significantly reduce its attribution. We also tried larger values of $s^\text{offset}$ for attention-offset but found it lead to performance degradation.  This analysis shows that despite the small number of parameters associated with the focus vectors, they are able to effectively steer the model's focus. We provide a simple visualization of the trained focus-vector parameters in Figure \ref{fig:attr_analysis} (Appendix \ref{ref:app_aux_sample}).

\vspace{-1mm}
\paragraph{Ablation Studies} Table \ref{tab:attn_variant} shows several variants of focus vector on CNN/Dailymail. We first tune the hyper-parameters of focus vector only with the original dev set with $\bc^\text{attr}$, instead of human-annotated highlights. Despite this discrepancy, focus vector still achieves strong performance on the test set. This result shows that the use of human-annotated dev set is not crucial for our framework.
We then conduct an ablation study where we only apply focus vector on the first or last layer of the encoder, which reduces the number of parameters. We find that this results in marginal performance degradation. 
Finally, we jointly finetune focus vector and the whole model with the same loss function (Equation \ref{eq:attvec_loss}), where a separate and smaller learning rate is used for the model. Interestingly, the gain from model finetuning is very limited, which demonstrates the effectiveness of focus vector.

\begin{table}[]
\footnotesize
\addtolength{\tabcolsep}{-3.1pt}
\centering
\begin{tabular}{@{}c|ccc@{}}
\toprule
\multirow{2}{*}{\textbf{Model}} & \multicolumn{3}{c}{\textbf{CNN/Dailymail}}               \\ 
                                & \textbf{PPL} & \textbf{ROUGE-1/2/L} & \textbf{BERTScore} \\ \midrule
\multicolumn{1}{c|}{all-layer*}  & 4.48         & 45.92/23.03/32.98    & 89.78              \\ 
\multicolumn{1}{c|}{\begin{tabular}[c]{@{}c@{}}ori-dev with $\bc^\text{attr}$ \\ \end{tabular}} & 4.50  & 46.41/22.69/32.48 & 89.62 \\ \midrule
\multicolumn{1}{c|}{only first layer}  & 4.48         & 45.67/22.63/32.45    & 89.59              \\
\multicolumn{1}{c|}{only last layer} & 4.48         & 46.06/22.84/32.69    & 89.69              \\ \midrule
\multicolumn{1}{c|}{plus model finetune}    & 4.49 & 46.65/23.54/33.30 & 89.82 \\ \bottomrule
\end{tabular}
\vspace{-1mm}
\caption{Performance of different variants of focus-vector trained on CNN/Dailymail. all-layer* refers to our proposed modelling (also reported in Table \ref{tab:main_res}).}
\vspace{-1mm}
\label{tab:attn_variant}
\end{table}

%% file: sec_related.tex
\vspace{-1mm}
\section{Related Work}
\vspace{-1mm}

Our proposed focus-vector framework is closely related to the research topics of controllable text generation, LM adaptation, and attention/attribution analysis, which we review below. 

\vspace{-1mm}
\paragraph{Controllable Text Generation} 


Prior work on controllable summarization introduced various types of control mechanisms. \citet{fan2017controllable, saito2020abstractive} extract entity, keyword or length, as additional supervision during training. \citet{DBLP:journals/corr/abs-1808-10792} trains a token-level content selection module, where the supervision is by aligning the summaries to the documents. \citep{song2021new} proposes a two-staged generation strategy and \citet{goyal2021hydrasum} incorporates multiple decoders into a transformer framework. Some recent work \citep{he2020ctrlsum, dou2020gsum} uses prompts to control the generation. Lexically constrained decoding \citep{post-vilar-2018-fast} has also been used to enforce certain key phrases to be included in the summary \citep{yuning2020constrainedsummary}.

Existing work on controllable dialogue response generation include using conditional variational autoencoders~\citep{zhao2017learning, li2020optimus}, and incorporating external knowledge into the conversational agent using knowledge graphs \citep{cui2021knowledge,moon2019opendialkg}, unstructured documents \citep{kim2020sequential}, or dialogue context \citep{zhao2020knowledge}. There is also a line of work on promoting the diversity or consistency of the model \citep{diversityjiwei16,negtrain19tianxing,li-etal-2020-dont}.

In open-ended language generation, a series of approaches have been proposed to control for some attribute (e.g., topic) of the generation \citep{nitish2019ctrl,Dathathri2020Plug,ben2020gedi,yang-klein-2021-fudge}. Some of these studies utilize a trained classifier to guide the generative model towards the desired attribute.

\vspace{-1mm}
\paragraph{LM Adaptation} Our proposed focus vector framework is also inspired by a series of recent works on prompting or light-weight LM adaptation. \citet{xiang2021prefixtuning}, followed by \citet{brian2021scaleprompt} and \citet{DBLP:journals/corr/abs-2104-05240}, propose \textit{prefix tuning}, where continuous task-specific input vectors are tuned to adapt the pretrained LM to a down-stream task with supervised data, and the model is kept fixed. 

There is also a line of works on \textit{adapter-tuning}, which  insert and finetune task-specific layers (adapters) between each layer of the pretrained LM \citep{pmlr-v97-houlsby19a,lin-etal-2020-exploring,pfeiffer-etal-2021-adapterfusion}. More recently, \citet{guo-etal-2021-parameter} and \citet{BenZaken2020BitFitSP} propose to finetune only a small subset of a pretrained model's parameters, and achieves strong performance on GLUE or other tasks \citep{wang-etal-2018-glue,tianxing21knowledgeprobe}. 

\vspace{-1mm}
\paragraph{Attention Analysis and Attribution Methods} Due to the ubiquity of the attention module in current NLP models, various work has studied how the module captures linguistic phenomena in the input \citep{clark-etal-2019-bert,kovaleva-etal-2019-revealing,kobayashi-etal-2020-attention}. It has also been used as a tool to interpret the model's predictions \citep{wang-etal-2016-attention, lee-etal-2017-interactive, ghaeini-etal-2018-interpreting}.

Recently, there have been a series of studies discussing the use of attention weights for interpretability  \citep{sarthak19attnotexp,wiegreffe-pinter-2019-attention,bastings-filippova-2020-elephant,DBLP:journals/corr/abs-1906-03731}, and it has been argued that attribution methods are a better choice to explain the model's predictions. The poor alignment performance of attention weights that we get in Table \ref{tab:attribute_res}, on some level, are in agreement with that argument. Our work is also related to the line of work on interpreting black box models through \emph{rationales} \citep{lei-etal-2016-rationalizing,bastings-etal-2019-interpretable}, which are typically (discrete) subsets of the input that are used to predict the output.
Finally, several recent works \citep{xu-durrett-2021-dissecting,ding-koehn-2021-evaluating} have compared different attribution methods for interpreting NLP models.


In comparison to the aforementioned works, our major innovations are two fold: (1) Our goal is to control the \textit{focus} of pretrained models, and thereby steer the model's generation, and our proposed focus vectors are compatible with the standard transformer architecture; (2) We utilize attribution methods to obtain automatic annotations for focus-vector training. Therefore, our framework can be applied to a wide range of NLG applications.

%% file: appendix.tex
\section*{Appendices}

\section{End-to-end Datasets}
\label{sec:app_datasets}

The PersonaChat dataset contains 8,939 dialogues for training, 1,000 for validation, and 968 for test. For each turn in the dialogue, we concatenate the persona of the speaker and the dialogue history as input, and train the base model to generate the current utterance. In some cases, the dialogue history is long and exceeds the input limit of the model, in which case we truncate the dialogue at the sentence level. The average number of sentences is around 11 after truncation.

The CNN/Dailymail dataset contains 287,113 training examples, and 13,368 / 11,490 examples for validation / test. We apply the same truncation strategy as \textit{PersonaChat} during preprocessing. The processed articles have an average length of 748 tokens, and the reference summaries have an average length of 67 tokens.

\section{Human-annotated Evaluation Data Collection}
\label{sec:app_eval}

\begin{figure}[]
    \centering
    \includegraphics[width=0.95\linewidth]{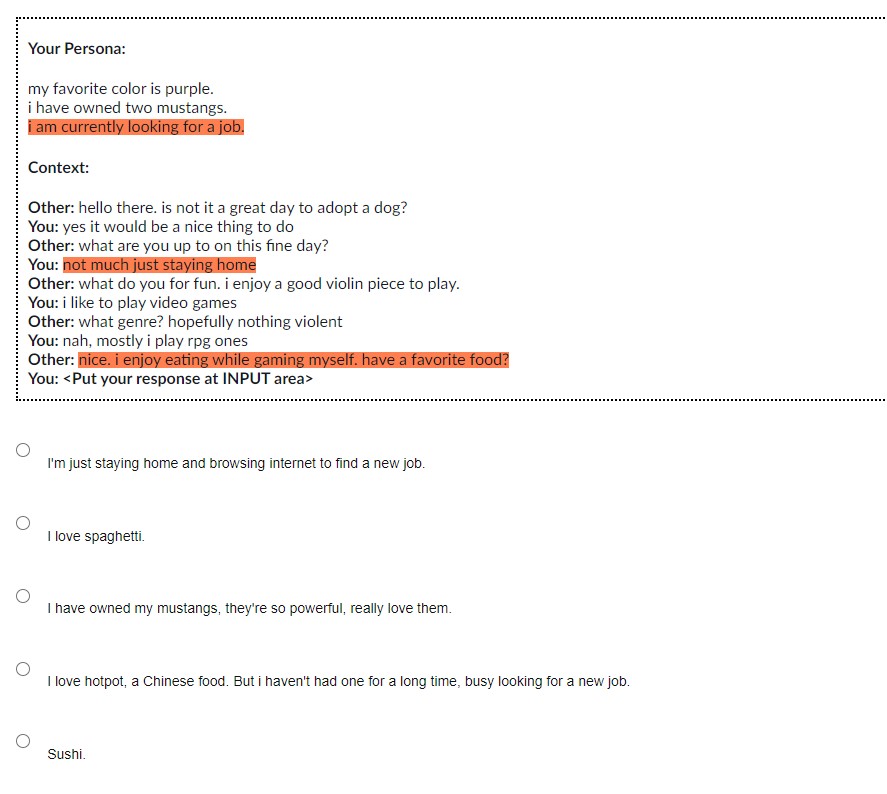}
    \caption{An example of our AMT qualification test for PersonaChat. We have chosen the highlights in the context, and the turker is supposed to choose a response that not only continues the dialogue, but also is relevant to the highlights.}
    \vspace{-0.5cm}
    \label{fig:qualification}
\end{figure}

To improve the quality of collected dataset, we design a qualification test, which the turkers need to pass before they can work on real assignments.  The test is designed to help turkers understand our task better. For PersonaChat, we give turkers two dialogue samples with pre-selected highlights, and ask them to choose the appropriate response that not only continues the dialogue, but also is relevant to the highlights. For CNN/Dailymail , the turkers are shown two example articles and the corresponding reference summaries. We have already picked some highlights in the article, but there is one highlight missing. And the turker is required to pick the missing highlight. The interface for the PersonaChat qualification test is shown in Figure \ref{fig:qualification}.

\begin{figure*}[]
    \centering
    \includegraphics[width=0.95\linewidth]{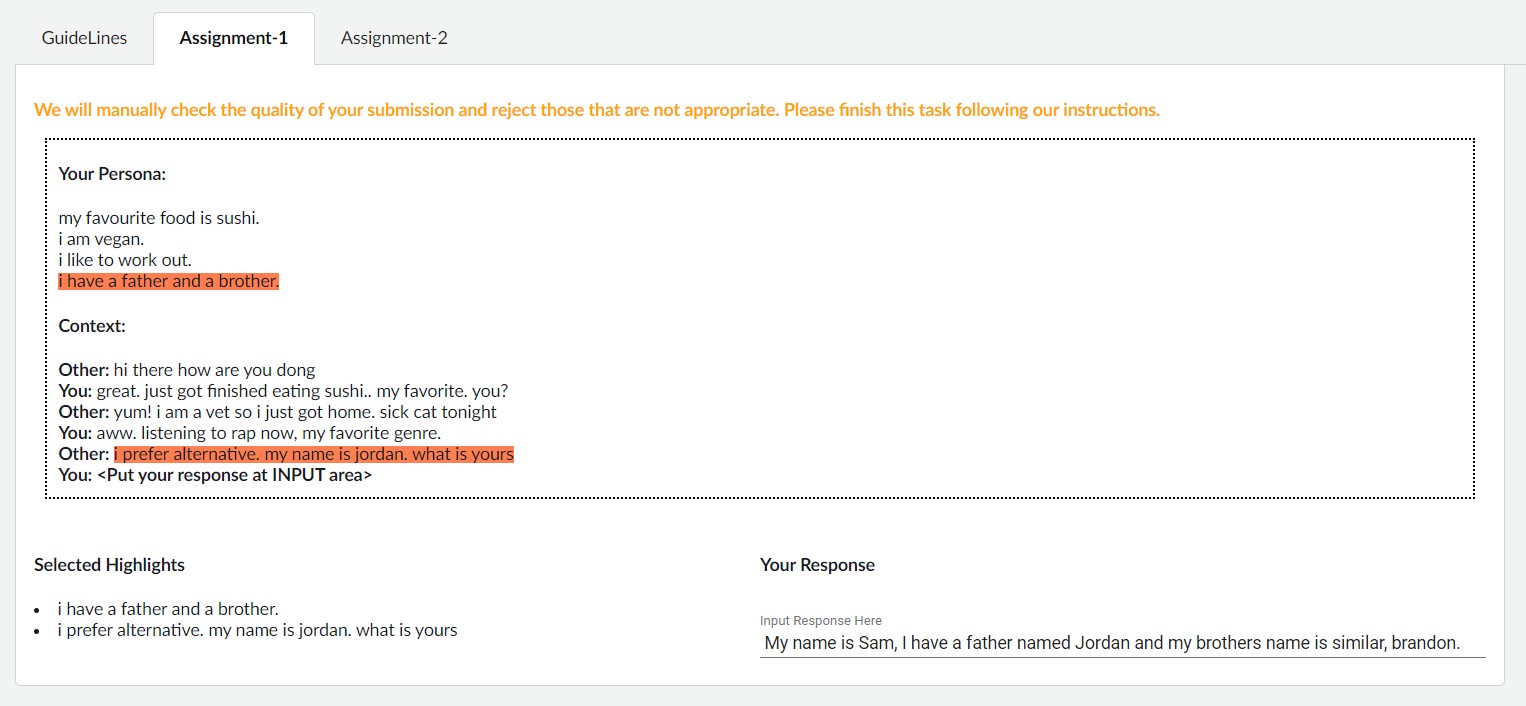}
    \caption{An example of our AMT interface on PersonaChat. The highlights and a response are labeled by a turker.}
    \label{fig:sample_interface}
\end{figure*}

We also add multiple checks in our script to prevent trivial answers. We ban trivial copy\&paste from the given context. A time check is added that requires turker to spend at least 60 seconds on a single HIT. For the two assignments in PersonaChat, we add a content check that prevents duplicate highlights or response. We show our interface for PersonaChat in Figure \ref{fig:sample_interface}. Despite these checks and the qualification tests, there still exist a small number of misbehaving turkers who attempt to cheat. Therefore we also manually monitor the incoming submissions, and ban misbehaving turkers and filter out their submissions.

More examples of our interface and instructions can be found in our uploaded data samples.

\section{Implementation Details}
\label{sec:app_implement}

For the attention-offset baseline, we tune the offset $s^\text{offset}$ in a  fine-grained manner, on the human-annotated dev set. We first set a relatively large max value (100) and get 20 evenly spaced numbers inside the interval $(0, 100)$. Then we calculate model PPL on the dev set with $s^\text{offset}$ set to these different offsets. Then we do another search in the interval that has lowest PPL. We repeat this iteration multiple times, and stops when PPL change is smaller than $1e^{-3}$. The final tuned value for Blenderbot is around 3.02, and around 0.17 for BART.

\begin{figure}[]
    \centering
    \includegraphics[width=0.95\linewidth]{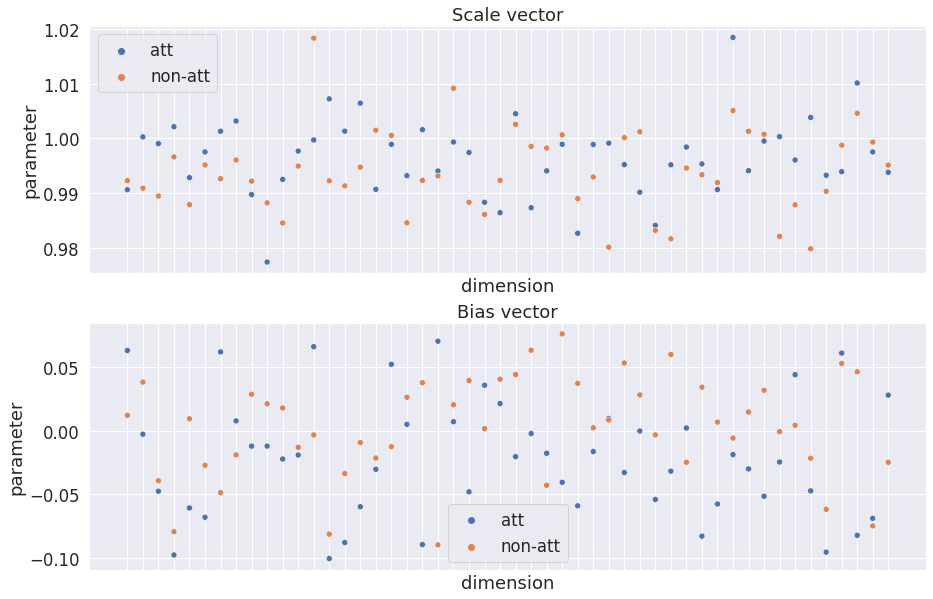}
    \caption{50 random dimensions of the trained focus vector on first encoder layer of the BART model.}
    \vspace{-0.5cm}
    \label{fig:att_param}
\end{figure}

\section{Auxiliary Results and Examples}
\label{ref:app_aux_sample}

In Figure \ref{fig:att_param}, we provide a simple visualization of the trained focus vectors of BART. To make the figure easy to grasp, we randomly sample 50 dimensions (out of 768) of trained focus-vector parameters. In the figure only the trained parameters for the first encoder layer is shown,  and we find that the patterns for the other layers are similar.

We first find that the learned scaling parameters are close to 1 and the bias parameters are close to zero. This implies that the transformation defined by the focus-vec is not drastic, and a relatively small tweak could be enough to steer the model's focus.

An interesting pattern can be observed for the scaling vectors.  $\theta_\text{scale-nonfocus}$ tends to down-scale the embedding, comparing to $\theta_\text{scale-focus}$. This matches our intuition that the embeddings on the non-highlighted positions should be down-played.

\begin{table*}[]
\footnotesize
\centering
\begin{tabular}{p{0.05\linewidth}p{0.4\linewidth}|p{0.4\linewidth}}
\toprule
\multicolumn{1}{r|}{persona:} &
  My parents did not want me. \textbf{It's a dead end job so i am looking for something different.} I was raised by my uncle and aunt. &
  My parents did not want me. It's a dead end job so i am looking for something different. \textbf{I was raised by my uncle and aunt.} \\ \midrule
\multicolumn{1}{r|}{context:} &
  \begin{tabular}[c]{@{}l@{}}{[}P-1{]}:  Hello how are you doing?\\ {[}P-2{]}:  Pretty good. how are you?\\ {[}P-1{]}:  I am making dinner getting ready for bed.\\ {[}P-2{]}:  I just got off work.\\ {[}P-1{]}:  \textbf{I am looking for something different to do.}\\ {[}P-2{]}:  {[}TO ANSWER{]}\end{tabular} &
  \begin{tabular}[c]{@{}l@{}}{[}P-1{]}:  Hello how are you doing?\\ {[}P-2{]}:  Pretty good. how are you?\\ {[}P-1{]}:  I am making dinner getting ready for bed.\\ {[}P-2{]}:  I just got off work.\\ {[}P-1{]}:  \textbf{I am looking for something different to do.}\\ {[}P-2{]}:  {[}TO ANSWER{]}\end{tabular} \\ \midrule
  \multicolumn{1}{r|}{reference:} &
  Do you know anywhere that's hiring work from home? I'm \textcolor{red}{looking for a new job.} &
  I grow up with \textcolor{red}{my aunt} and we always had scripture study after dinner, you can try it. \\
  \midrule
\multicolumn{1}{r|}{vanilla:} &
  My parents wanted me to be a doctor like them, but i do not want to do that. &
  My parents wanted me to be a doctor like them, but i do not want to do that. \\
\multicolumn{1}{r|}{keyword:} &
  What do you do for \textcolor{red}{work}? I am a cashier at a fast food restaurant. &
  What do you do for work? I am a cashier at a fast food restaurant. \\
\multicolumn{1}{r|}{att-offset:} &
  What do you do for a living? I have been \textcolor{red}{looking for a new job} myself. &
  Do you have any relatives you can ask for help with? my family was always there for me. \\
\multicolumn{1}{r|}{focus-vec:} &
  What do you do for a living? I am unemployed right now \textcolor{red}{looking for a job}. &
  What do you do for a living? I work with \textcolor{red}{my aunt} at her law firm. \\ \bottomrule
\end{tabular}
\caption{Generation samples from different attention control approaches for PersonaChat. The inputs on both sides are identical, except a different piece of persona is highlighted, which leads to different generations from att-offset and focus-vec.}
\label{tab:persona_sample2}
\end{table*}

\begin{table*}[]
\footnotesize
\centering
\begin{tabular}{p{0.05\linewidth}|p{0.85\linewidth}}
\toprule
\multicolumn{1}{r|}{article:} &
  England World Cup hopeful Elliot Daly has been named the Aviva Premiership’s player of the month for March$.^{[1]}$ \textbf{The uncapped Wasps centre, 22, scored two tries in March as the Coventry-based club maintained on course for a Champions Cup spot next season}$.^{[2]}$ \textbf{Equally comfortable at full back, Daly's form at centre has him pushing incumbent England No 13 Jonathan Joseph hard for his place}$.^{[3]}$ Elliot Daly is handed his player of the month award by Wasps director of rugby Dai Young$.^{[4]}$ He looks certain to be included in Stuart Lancaster's England's World Cup training squad named next month$.^{[5]}$\textbf{ Wasps director of rugby Dai Young said: 'If England are going to go out and win the Rugby World Cup, he’s the type of player you’ve got to have in the team so I’d certainly like to think he’s pushing for a place in the squad.}'$.^{[6]}$ BT Sport analyst Austin Healey added: 'For a good couple of years now us rugby watchers have known of the potential of Elliot Daly.'$^{[7]}$ $\cdots$ (LEAVE OUT UNIMPORTANT SENTECES) \\ \midrule
\multicolumn{1}{r|}{reference:} &
  Elliot Daly was in fine form at outside centre for Wasps in March. Daly, 22, has not yet been capped by England. \textcolor{red}{Wasps head coach Dai Young says England must pick him for World Cup.} \\ \midrule
\multicolumn{1}{r|}{vanilla:} &
  England World Cup hopeful Elliot Daly has been named Aviva Premiership's player of the month for March. The 22-year-old scored two tries in March as Coventry-based club maintained on course for a Champions Cup spot next season. Daly's form at centre has him pushing incumbent England No 13 Jonathan Joseph hard for his place. \\ \midrule
\multicolumn{1}{r|}{keyword:} &
 England World Cup hopeful Elliot Daly has been named Aviva Premiership's player of the month. The 22-year-old scored two tries in March as Coventry-based club maintained on course for a Champions Cup spot next season. Daly looks certain to be included in Stuart Lancaster's England's World Cup training squad.
\\  \midrule
\multicolumn{1}{r|}{att-offset:} &
England World Cup hopeful Elliot Daly has been named Aviva Premiership's player of the month for March. The uncapped Wasps centre scored two tries in March as Coventry-based club maintained on course for a Champions Cup spot next season. Daly's form at centre has him pushing incumbent England No 13 Jonathan Joseph hard for his place.  \\ \midrule
\multicolumn{1}{r|}{focus-vec:} &
  The uncapped Wasps centre scored two tries in March. Daly's form at centre has him pushing incumbent England No 13 Jonathan Joseph hard for his place. Wasps director of rugby Dai Young said: 'If England are going to go out and win the Rugby World Cup, \textcolor{red}{he's the type of player you’ve got to have in the team.'}\\ \bottomrule
\end{tabular}
\caption{Generation samples of different attention control approaches on CNN/Dailymail dataset. The highlighted setences are marked in bold. The span marked by red in the reference summary is captured by the generation from focus-vec, but not by att-offset. The sentence number marked in the input corresponds the sentence index in Figure \ref{fig:attr_analysis}. }
\label{tab:dailymail_sample}
\end{table*}

\begin{table*}[]
\footnotesize
\centering
\begin{tabular}{p{0.05\linewidth}p{0.85\linewidth}}
\toprule
\multicolumn{1}{r|}{article:} &
  \textbf{Kabul, Afghanistan (CNN)A suicide bomber detonated his explosives near a group of protesters in eastern Afghanistan on Thursday, killing 17 people and wounding dozens more, police said}$.^{[1]}$ \textbf{"An Afghan lawmaker taking part in the protests in the city of Khost was among the 64 people wounded,"$^{[2]}$ said Faizullah Ghairat}, \textbf{the provincial police chief Taliban spokesman Zabiullah Mujahid denied his group was responsible for the attack}$.^{[3]}$ No other organization has so far claimed responsibility$.^{[4]}$ Humayoon Humayoon, an Afghan member of parliament for Khost province, and the other protesters were on their way to join a larger rally against the provincial governor, according to Zahir Jan, an eyewitness$.^{[5]}$ The suicide attack hit the group around 10 a.m. local time, police said$.^{[6]}$ $\cdots$ (LEAVE OUT UNIMPORTANT SENTENCES) \\ \midrule
\multicolumn{1}{r|}{reference:} &
  \textcolor{red}{An Afghan lawmaker is among 64 people wounded in the attack, police say.} Taliban spokesman denies his group was responsible for the attack. \\ \midrule
\multicolumn{1}{r|}{vanilla:} &
  A suicide bomber detonates his explosives near a group of protesters, killing 17 people and wounding dozens more. The Taliban spokesman denies his group was responsible for the attack. No other organization has so far claimed responsibility for the attack. The suicide attack hit the group around 10 a.m. local time. \\ \midrule
\multicolumn{1}{r|}{keyword:} &
  Taliban spokesman Zabiullah Mujahid denies his group was responsible. No other organization has so far claimed responsibility. The suicide attack hit the group around 10 a.m. local time. \\
   \midrule
\multicolumn{1}{r|}{att-offset:} &
A suicide bomber detonates his explosives near a group of protesters, killing 17 people and wounding dozens more. The Taliban spokesman denies his group was responsible for the attack. No other organization has so far claimed responsibility for the attack. The suicide attack hit the group around 10 a.m. local time. \\ \midrule
\multicolumn{1}{r|}{focus-vec:} &
A suicide bomber detonates his explosives near a group of protesters, killing 17 people. \textcolor{red}{An Afghan lawmaker is among the 64 people wounded, police say.} Taliban spokesman Zabiullah Mujahid denies his group was responsible for the attack. No other organization has so far claimed responsibility. \\
   \bottomrule
\end{tabular}
\caption{Generation samples of different attention control approaches on CNN/Dailymail dataset. The span marked by red in the reference summary is captured by the generation from focus-vec, but not by att-offset.}
\label{tab:dailymail_sample2}
\end{table*}

%% file: main_acl.bbl
\begin{thebibliography}{72}
\expandafter\ifx\csname natexlab\endcsname\relax\def\natexlab#1{#1}\fi

\bibitem[{Adebayo et~al.(2018)Adebayo, Gilmer, Muelly, Goodfellow, Hardt, and
  Kim}]{NEURIPS2018_294a8ed2}
Julius Adebayo, Justin Gilmer, Michael Muelly, Ian Goodfellow, Moritz Hardt,
  and Been Kim. 2018.
\newblock \href
  {https://proceedings.neurips.cc/paper/2018/file/294a8ed24b1ad22ec2e7efea049b8737-Paper.pdf}
  {Sanity checks for saliency maps}.
\newblock In \emph{Advances in Neural Information Processing Systems},
  volume~31. Curran Associates, Inc.

\bibitem[{Baehrens et~al.(2010)Baehrens, Schroeter, Harmeling, Kawanabe,
  Hansen, and M{{\"u}}ller}]{JMLR:v11:baehrens10a}
David Baehrens, Timon Schroeter, Stefan Harmeling, Motoaki Kawanabe, Katja
  Hansen, and Klaus-Robert M{{\"u}}ller. 2010.
\newblock \href {http://jmlr.org/papers/v11/baehrens10a.html} {How to explain
  individual classification decisions}.
\newblock \emph{Journal of Machine Learning Research}, 11(61):1803--1831.

\bibitem[{Bahdanau et~al.(2016)Bahdanau, Cho, and Bengio}]{bahdanau2016neural}
Dzmitry Bahdanau, Kyunghyun Cho, and Yoshua Bengio. 2016.
\newblock \href {http://arxiv.org/abs/1409.0473} {Neural machine translation by
  jointly learning to align and translate}.

\bibitem[{Bastings et~al.(2019)Bastings, Aziz, and
  Titov}]{bastings-etal-2019-interpretable}
Jasmijn Bastings, Wilker Aziz, and Ivan Titov. 2019.
\newblock \href {https://doi.org/10.18653/v1/P19-1284} {Interpretable neural
  predictions with differentiable binary variables}.
\newblock In \emph{Proceedings of the 57th Annual Meeting of the Association
  for Computational Linguistics}, pages 2963--2977, Florence, Italy.
  Association for Computational Linguistics.

\bibitem[{Bastings and Filippova(2020)}]{bastings-filippova-2020-elephant}
Jasmijn Bastings and Katja Filippova. 2020.
\newblock \href {https://doi.org/10.18653/v1/2020.blackboxnlp-1.14} {The
  elephant in the interpretability room: Why use attention as explanation when
  we have saliency methods?}
\newblock In \emph{Proceedings of the Third BlackboxNLP Workshop on Analyzing
  and Interpreting Neural Networks for NLP}, pages 149--155, Online.
  Association for Computational Linguistics.

\bibitem[{Ben-Zaken et~al.(2020)Ben-Zaken, Ravfogel, and
  Goldberg}]{BenZaken2020BitFitSP}
Elad Ben-Zaken, Shauli Ravfogel, and Yoav Goldberg. 2020.
\newblock Bitfit: Simple parameter-efficient fine-tuning for transformer-based
  masked language-models.

\bibitem[{Brunner et~al.(2020)Brunner, Liu, Pascual, Richter, Ciaramita, and
  Wattenhofer}]{Brunner2020On}
Gino Brunner, Yang Liu, Damian Pascual, Oliver Richter, Massimiliano Ciaramita,
  and Roger Wattenhofer. 2020.
\newblock \href {https://openreview.net/forum?id=BJg1f6EFDB} {On
  identifiability in transformers}.
\newblock In \emph{International Conference on Learning Representations}.

\bibitem[{Campos et~al.(2020)Campos, Mangaravite, Pasquali, Jorge, Nunes, and
  Jatowt}]{campos2020yake}
Ricardo Campos, V{\'\i}tor Mangaravite, Arian Pasquali, Al{\'\i}pio Jorge,
  C{\'e}lia Nunes, and Adam Jatowt. 2020.
\newblock Yake! keyword extraction from single documents using multiple local
  features.
\newblock \emph{Information Sciences}, 509:257--289.

\bibitem[{Clark et~al.(2019)Clark, Khandelwal, Levy, and
  Manning}]{clark-etal-2019-bert}
Kevin Clark, Urvashi Khandelwal, Omer Levy, and Christopher~D. Manning. 2019.
\newblock \href {https://doi.org/10.18653/v1/W19-4828} {What does {BERT} look
  at? an analysis of {BERT}{'}s attention}.
\newblock In \emph{Proceedings of the 2019 ACL Workshop BlackboxNLP: Analyzing
  and Interpreting Neural Networks for NLP}, pages 276--286, Florence, Italy.
  Association for Computational Linguistics.

\bibitem[{Cui et~al.(2021)Cui, Wu, Liu, and Zhang}]{cui2021knowledge}
Leyang Cui, Yu~Wu, Shujie Liu, and Yue Zhang. 2021.
\newblock Knowledge enhanced fine-tuning for better handling unseen entities in
  dialogue generation.
\newblock \emph{arXiv preprint arXiv:2109.05487}.

\bibitem[{Dathathri et~al.(2020)Dathathri, Madotto, Lan, Hung, Frank, Molino,
  Yosinski, and Liu}]{Dathathri2020Plug}
Sumanth Dathathri, Andrea Madotto, Janice Lan, Jane Hung, Eric Frank, Piero
  Molino, Jason Yosinski, and Rosanne Liu. 2020.
\newblock \href {https://openreview.net/forum?id=H1edEyBKDS} {Plug and play
  language models: A simple approach to controlled text generation}.
\newblock In \emph{International Conference on Learning Representations}.

\bibitem[{Denil et~al.(2014)Denil, Demiraj, and
  de~Freitas}]{DBLP:journals/corr/DenilDF14}
Misha Denil, Alban Demiraj, and Nando de~Freitas. 2014.
\newblock \href {http://arxiv.org/abs/1412.6815} {Extraction of salient
  sentences from labelled documents}.
\newblock \emph{CoRR}, abs/1412.6815.

\bibitem[{Ding and Koehn(2021)}]{ding-koehn-2021-evaluating}
Shuoyang Ding and Philipp Koehn. 2021.
\newblock \href {https://doi.org/10.18653/v1/2021.naacl-main.399} {Evaluating
  saliency methods for neural language models}.
\newblock In \emph{Proceedings of the 2021 Conference of the North American
  Chapter of the Association for Computational Linguistics: Human Language
  Technologies}, pages 5034--5052, Online. Association for Computational
  Linguistics.

\bibitem[{Dong et~al.(2021)Dong, Bhagavatula, Lu, Hwang, Bosselut, Cheung, and
  Choi}]{dong-etal-2021-fly}
Yue Dong, Chandra Bhagavatula, Ximing Lu, Jena~D. Hwang, Antoine Bosselut,
  Jackie Chi~Kit Cheung, and Yejin Choi. 2021.
\newblock \href {https://doi.org/10.18653/v1/2021.findings-acl.107} {On-the-fly
  attention modulation for neural generation}.
\newblock In \emph{Findings of the Association for Computational Linguistics:
  ACL-IJCNLP 2021}, pages 1261--1274, Online. Association for Computational
  Linguistics.

\bibitem[{Dou et~al.(2020)Dou, Liu, Hayashi, Jiang, and Neubig}]{dou2020gsum}
Zi-Yi Dou, Pengfei Liu, Hiroaki Hayashi, Zhengbao Jiang, and Graham Neubig.
  2020.
\newblock Gsum: A general framework for guided neural abstractive
  summarization.
\newblock \emph{arXiv preprint arXiv:2010.08014}.

\bibitem[{Fan et~al.(2017)Fan, Grangier, and Auli}]{fan2017controllable}
Angela Fan, David Grangier, and Michael Auli. 2017.
\newblock Controllable abstractive summarization.
\newblock \emph{arXiv preprint arXiv:1711.05217}.

\bibitem[{Gehrmann et~al.(2018)Gehrmann, Deng, and
  Rush}]{DBLP:journals/corr/abs-1808-10792}
Sebastian Gehrmann, Yuntian Deng, and Alexander~M. Rush. 2018.
\newblock \href {http://arxiv.org/abs/1808.10792} {Bottom-up abstractive
  summarization}.
\newblock \emph{CoRR}, abs/1808.10792.

\bibitem[{Ghaeini et~al.(2018)Ghaeini, Fern, and
  Tadepalli}]{ghaeini-etal-2018-interpreting}
Reza Ghaeini, Xiaoli Fern, and Prasad Tadepalli. 2018.
\newblock \href {https://doi.org/10.18653/v1/D18-1537} {Interpreting recurrent
  and attention-based neural models: a case study on natural language
  inference}.
\newblock In \emph{Proceedings of the 2018 Conference on Empirical Methods in
  Natural Language Processing}, pages 4952--4957, Brussels, Belgium.
  Association for Computational Linguistics.

\bibitem[{Goyal et~al.(2021)Goyal, Rajani, Liu, and
  Kry{\'s}ci{\'n}ski}]{goyal2021hydrasum}
Tanya Goyal, Nazneen~Fatema Rajani, Wenhao Liu, and Wojciech
  Kry{\'s}ci{\'n}ski. 2021.
\newblock Hydrasum--disentangling stylistic features in text summarization
  using multi-decoder models.
\newblock \emph{arXiv preprint arXiv:2110.04400}.

\bibitem[{Guo et~al.(2021)Guo, Rush, and Kim}]{guo-etal-2021-parameter}
Demi Guo, Alexander Rush, and Yoon Kim. 2021.
\newblock \href {https://doi.org/10.18653/v1/2021.acl-long.378}
  {Parameter-efficient transfer learning with diff pruning}.
\newblock In \emph{Proceedings of the 59th Annual Meeting of the Association
  for Computational Linguistics and the 11th International Joint Conference on
  Natural Language Processing (Volume 1: Long Papers)}, pages 4884--4896,
  Online. Association for Computational Linguistics.

\bibitem[{He et~al.(2020)He, Kry{\'s}ci{\'n}ski, McCann, Rajani, and
  Xiong}]{he2020ctrlsum}
Junxian He, Wojciech Kry{\'s}ci{\'n}ski, Bryan McCann, Nazneen Rajani, and
  Caiming Xiong. 2020.
\newblock Ctrlsum: Towards generic controllable text summarization.
\newblock \emph{arXiv preprint arXiv:2012.04281}.

\bibitem[{He et~al.(2021)He, Cho, and Glass}]{tianxing21knowledgeprobe}
Tianxing He, Kyunghyun Cho, and James~R. Glass. 2021.
\newblock \href {http://arxiv.org/abs/2109.02772} {An empirical study on
  few-shot knowledge probing for pretrained language models}.
\newblock \emph{CoRR}, abs/2109.02772.

\bibitem[{He and Glass(2019)}]{negtrain19tianxing}
Tianxing He and James~R. Glass. 2019.
\newblock \href {http://arxiv.org/abs/1903.02134} {Negative training for neural
  dialogue response generation}.
\newblock \emph{CoRR}, abs/1903.02134.

\bibitem[{Hermann et~al.(2015)Hermann, Kocisky, Grefenstette, Espeholt, Kay,
  Suleyman, and Blunsom}]{hermann2015teaching}
Karl~Moritz Hermann, Tomas Kocisky, Edward Grefenstette, Lasse Espeholt, Will
  Kay, Mustafa Suleyman, and Phil Blunsom. 2015.
\newblock Teaching machines to read and comprehend.
\newblock \emph{Advances in neural information processing systems},
  28:1693--1701.

\bibitem[{Holtzman et~al.(2019)Holtzman, Buys, Forbes, and Choi}]{curious19ari}
Ari Holtzman, Jan Buys, Maxwell Forbes, and Yejin Choi. 2019.
\newblock \href {http://arxiv.org/abs/1904.09751} {The curious case of neural
  text degeneration}.
\newblock \emph{CoRR}, abs/1904.09751.

\bibitem[{Houlsby et~al.(2019)Houlsby, Giurgiu, Jastrzebski, Morrone,
  De~Laroussilhe, Gesmundo, Attariyan, and Gelly}]{pmlr-v97-houlsby19a}
Neil Houlsby, Andrei Giurgiu, Stanislaw Jastrzebski, Bruna Morrone, Quentin
  De~Laroussilhe, Andrea Gesmundo, Mona Attariyan, and Sylvain Gelly. 2019.
\newblock \href {https://proceedings.mlr.press/v97/houlsby19a.html}
  {Parameter-efficient transfer learning for {NLP}}.
\newblock In \emph{Proceedings of the 36th International Conference on Machine
  Learning}, volume~97 of \emph{Proceedings of Machine Learning Research},
  pages 2790--2799. PMLR.

\bibitem[{Huang et~al.(2015)Huang, Xu, and Yu}]{DBLP:journals/corr/HuangXY15}
Zhiheng Huang, Wei Xu, and Kai Yu. 2015.
\newblock \href {http://arxiv.org/abs/1508.01991} {Bidirectional {LSTM-CRF}
  models for sequence tagging}.
\newblock \emph{CoRR}, abs/1508.01991.

\bibitem[{Jain and Wallace(2019)}]{sarthak19attnotexp}
Sarthak Jain and Byron~C. Wallace. 2019.
\newblock \href {http://arxiv.org/abs/1902.10186} {Attention is not
  explanation}.
\newblock \emph{CoRR}, abs/1902.10186.

\bibitem[{Keskar et~al.(2019)Keskar, McCann, Varshney, Xiong, and
  Socher}]{nitish2019ctrl}
Nitish~Shirish Keskar, Bryan McCann, Lav~R. Varshney, Caiming Xiong, and
  Richard Socher. 2019.
\newblock \href {http://arxiv.org/abs/1909.05858} {{CTRL:} {A} conditional
  transformer language model for controllable generation}.
\newblock \emph{CoRR}, abs/1909.05858.

\bibitem[{Kim et~al.(2020)Kim, Ahn, and Kim}]{kim2020sequential}
Byeongchang Kim, Jaewoo Ahn, and Gunhee Kim. 2020.
\newblock Sequential latent knowledge selection for knowledge-grounded
  dialogue.
\newblock \emph{arXiv preprint arXiv:2002.07510}.

\bibitem[{Kingma and Ba(2014)}]{kingma2014adam}
Diederik~P Kingma and Jimmy Ba. 2014.
\newblock Adam: A method for stochastic optimization.
\newblock \emph{arXiv preprint arXiv:1412.6980}.

\bibitem[{Kobayashi et~al.(2020)Kobayashi, Kuribayashi, Yokoi, and
  Inui}]{kobayashi-etal-2020-attention}
Goro Kobayashi, Tatsuki Kuribayashi, Sho Yokoi, and Kentaro Inui. 2020.
\newblock \href {https://doi.org/10.18653/v1/2020.emnlp-main.574} {Attention is
  not only a weight: Analyzing transformers with vector norms}.
\newblock In \emph{Proceedings of the 2020 Conference on Empirical Methods in
  Natural Language Processing (EMNLP)}, pages 7057--7075, Online. Association
  for Computational Linguistics.

\bibitem[{Kovaleva et~al.(2019)Kovaleva, Romanov, Rogers, and
  Rumshisky}]{kovaleva-etal-2019-revealing}
Olga Kovaleva, Alexey Romanov, Anna Rogers, and Anna Rumshisky. 2019.
\newblock \href {https://doi.org/10.18653/v1/D19-1445} {Revealing the dark
  secrets of {BERT}}.
\newblock In \emph{Proceedings of the 2019 Conference on Empirical Methods in
  Natural Language Processing and the 9th International Joint Conference on
  Natural Language Processing (EMNLP-IJCNLP)}, pages 4365--4374, Hong Kong,
  China. Association for Computational Linguistics.

\bibitem[{Krause et~al.(2020)Krause, Gotmare, McCann, Keskar, Joty, Socher, and
  Rajani}]{ben2020gedi}
Ben Krause, Akhilesh~Deepak Gotmare, Bryan McCann, Nitish~Shirish Keskar,
  Shafiq~R. Joty, Richard Socher, and Nazneen~Fatema Rajani. 2020.
\newblock \href {http://arxiv.org/abs/2009.06367} {Gedi: Generative
  discriminator guided sequence generation}.
\newblock \emph{CoRR}, abs/2009.06367.

\bibitem[{Lee et~al.(2017)Lee, Shin, and Kim}]{lee-etal-2017-interactive}
Jaesong Lee, Joong-Hwi Shin, and Jun-Seok Kim. 2017.
\newblock \href {https://doi.org/10.18653/v1/D17-2021} {Interactive
  visualization and manipulation of attention-based neural machine
  translation}.
\newblock In \emph{Proceedings of the 2017 Conference on Empirical Methods in
  Natural Language Processing: System Demonstrations}, pages 121--126,
  Copenhagen, Denmark. Association for Computational Linguistics.

\bibitem[{Lei et~al.(2016)Lei, Barzilay, and
  Jaakkola}]{lei-etal-2016-rationalizing}
Tao Lei, Regina Barzilay, and Tommi Jaakkola. 2016.
\newblock \href {https://doi.org/10.18653/v1/D16-1011} {Rationalizing neural
  predictions}.
\newblock In \emph{Proceedings of the 2016 Conference on Empirical Methods in
  Natural Language Processing}, pages 107--117, Austin, Texas. Association for
  Computational Linguistics.

\bibitem[{Lester et~al.(2021)Lester, Al{-}Rfou, and
  Constant}]{brian2021scaleprompt}
Brian Lester, Rami Al{-}Rfou, and Noah Constant. 2021.
\newblock \href {http://arxiv.org/abs/2104.08691} {The power of scale for
  parameter-efficient prompt tuning}.
\newblock \emph{CoRR}, abs/2104.08691.

\bibitem[{Lewis et~al.(2019)Lewis, Liu, Goyal, Ghazvininejad, Mohamed, Levy,
  Stoyanov, and Zettlemoyer}]{lewis2019bart}
Mike Lewis, Yinhan Liu, Naman Goyal, Marjan Ghazvininejad, Abdelrahman Mohamed,
  Omer Levy, Ves Stoyanov, and Luke Zettlemoyer. 2019.
\newblock Bart: Denoising sequence-to-sequence pre-training for natural
  language generation, translation, and comprehension.
\newblock \emph{arXiv preprint arXiv:1910.13461}.

\bibitem[{Li et~al.(2020{\natexlab{a}})Li, Gao, Li, Peng, Li, Zhang, and
  Gao}]{li2020optimus}
Chunyuan Li, Xiang Gao, Yuan Li, Baolin Peng, Xiujun Li, Yizhe Zhang, and
  Jianfeng Gao. 2020{\natexlab{a}}.
\newblock Optimus: Organizing sentences via pre-trained modeling of a latent
  space.
\newblock \emph{arXiv preprint arXiv:2004.04092}.

\bibitem[{Li et~al.(2016{\natexlab{a}})Li, Galley, Brockett, Gao, and
  Dolan}]{diversityjiwei16}
Jiwei Li, Michel Galley, Chris Brockett, Jianfeng Gao, and Bill Dolan.
  2016{\natexlab{a}}.
\newblock A diversity-promoting objective function for neural conversation
  models.
\newblock In \emph{{NAACL} {HLT} 2016, The 2016 Conference of the North
  American Chapter of the Association for Computational Linguistics: Human
  Language Technologies, San Diego California, USA, June 12-17, 2016}, pages
  110--119.

\bibitem[{Li et~al.(2016{\natexlab{b}})Li, Monroe, and
  Jurafsky}]{DBLP:journals/corr/LiMJ16a}
Jiwei Li, Will Monroe, and Dan Jurafsky. 2016{\natexlab{b}}.
\newblock \href {http://arxiv.org/abs/1612.08220} {Understanding neural
  networks through representation erasure}.
\newblock \emph{CoRR}, abs/1612.08220.

\bibitem[{Li et~al.(2020{\natexlab{b}})Li, Roller, Kulikov, Welleck, Boureau,
  Cho, and Weston}]{li-etal-2020-dont}
Margaret Li, Stephen Roller, Ilia Kulikov, Sean Welleck, Y-Lan Boureau,
  Kyunghyun Cho, and Jason Weston. 2020{\natexlab{b}}.
\newblock \href {https://doi.org/10.18653/v1/2020.acl-main.428} {Don{'}t say
  that! making inconsistent dialogue unlikely with unlikelihood training}.
\newblock In \emph{Proceedings of the 58th Annual Meeting of the Association
  for Computational Linguistics}, pages 4715--4728, Online. Association for
  Computational Linguistics.

\bibitem[{Li and Liang(2021)}]{xiang2021prefixtuning}
Xiang~Lisa Li and Percy Liang. 2021.
\newblock \href {http://arxiv.org/abs/2101.00190} {Prefix-tuning: Optimizing
  continuous prompts for generation}.
\newblock \emph{CoRR}, abs/2101.00190.

\bibitem[{Lin(2004)}]{lin2004rouge}
Chin-Yew Lin. 2004.
\newblock Rouge: A package for automatic evaluation of summaries.
\newblock In \emph{Text summarization branches out}, pages 74--81.

\bibitem[{Lin et~al.(2020)Lin, Madotto, and Fung}]{lin-etal-2020-exploring}
Zhaojiang Lin, Andrea Madotto, and Pascale Fung. 2020.
\newblock \href {https://doi.org/10.18653/v1/2020.findings-emnlp.41} {Exploring
  versatile generative language model via parameter-efficient transfer
  learning}.
\newblock In \emph{Findings of the Association for Computational Linguistics:
  EMNLP 2020}, pages 441--459, Online. Association for Computational
  Linguistics.

\bibitem[{Mao et~al.(2020)Mao, Ren, Ji, and Han}]{yuning2020constrainedsummary}
Yuning Mao, Xiang Ren, Heng Ji, and Jiawei Han. 2020.
\newblock \href {http://arxiv.org/abs/2010.12723} {Constrained abstractive
  summarization: Preserving factual consistency with constrained generation}.
\newblock \emph{CoRR}, abs/2010.12723.

\bibitem[{Moon et~al.(2019)Moon, Shah, Kumar, and Subba}]{moon2019opendialkg}
Seungwhan Moon, Pararth Shah, Anuj Kumar, and Rajen Subba. 2019.
\newblock Opendialkg: Explainable conversational reasoning with attention-based
  walks over knowledge graphs.
\newblock In \emph{Proceedings of the 57th Annual Meeting of the Association
  for Computational Linguistics}, pages 845--854.

\bibitem[{Nallapati et~al.(2016)Nallapati, Zhou, dos Santos, glar
  Gul{\c{c}}ehre, and Xiang}]{nallapati2016abstractive}
Ramesh Nallapati, Bowen Zhou, Cicero dos Santos, {\c{C}}a~glar Gul{\c{c}}ehre,
  and Bing Xiang. 2016.
\newblock Abstractive text summarization using sequence-to-sequence rnns and
  beyond.
\newblock \emph{CoNLL 2016}, page 280.

\bibitem[{Pfeiffer et~al.(2021)Pfeiffer, Kamath, R{\"u}ckl{\'e}, Cho, and
  Gurevych}]{pfeiffer-etal-2021-adapterfusion}
Jonas Pfeiffer, Aishwarya Kamath, Andreas R{\"u}ckl{\'e}, Kyunghyun Cho, and
  Iryna Gurevych. 2021.
\newblock \href {https://doi.org/10.18653/v1/2021.eacl-main.39}
  {{A}dapter{F}usion: Non-destructive task composition for transfer learning}.
\newblock In \emph{Proceedings of the 16th Conference of the European Chapter
  of the Association for Computational Linguistics: Main Volume}, pages
  487--503, Online. Association for Computational Linguistics.

\bibitem[{Post and Vilar(2018)}]{post-vilar-2018-fast}
Matt Post and David Vilar. 2018.
\newblock \href {https://doi.org/10.18653/v1/N18-1119} {Fast lexically
  constrained decoding with dynamic beam allocation for neural machine
  translation}.
\newblock In \emph{Proceedings of the 2018 Conference of the North {A}merican
  Chapter of the Association for Computational Linguistics: Human Language
  Technologies, Volume 1 (Long Papers)}, pages 1314--1324, New Orleans,
  Louisiana. Association for Computational Linguistics.

\bibitem[{Raffel et~al.(2020)Raffel, Shazeer, Roberts, Lee, Narang, Matena,
  Zhou, Li, and Liu}]{2020t5}
Colin Raffel, Noam Shazeer, Adam Roberts, Katherine Lee, Sharan Narang, Michael
  Matena, Yanqi Zhou, Wei Li, and Peter~J. Liu. 2020.
\newblock \href {http://jmlr.org/papers/v21/20-074.html} {Exploring the limits
  of transfer learning with a unified text-to-text transformer}.
\newblock \emph{Journal of Machine Learning Research}, 21(140):1--67.

\bibitem[{Roller et~al.(2020)Roller, Dinan, Goyal, Ju, Williamson, Liu, Xu,
  Ott, Shuster, Smith et~al.}]{roller2020recipes}
Stephen Roller, Emily Dinan, Naman Goyal, Da~Ju, Mary Williamson, Yinhan Liu,
  Jing Xu, Myle Ott, Kurt Shuster, Eric~M Smith, et~al. 2020.
\newblock Recipes for building an open-domain chatbot.
\newblock \emph{arXiv preprint arXiv:2004.13637}.

\bibitem[{Rush et~al.(2015)Rush, Chopra, and Weston}]{rush-etal-2015-neural}
Alexander~M. Rush, Sumit Chopra, and Jason Weston. 2015.
\newblock \href {https://doi.org/10.18653/v1/D15-1044} {A neural attention
  model for abstractive sentence summarization}.
\newblock In \emph{Proceedings of the 2015 Conference on Empirical Methods in
  Natural Language Processing}, pages 379--389, Lisbon, Portugal. Association
  for Computational Linguistics.

\bibitem[{Saito et~al.(2020)Saito, Nishida, Nishida, and
  Tomita}]{saito2020abstractive}
Itsumi Saito, Kyosuke Nishida, Kosuke Nishida, and Junji Tomita. 2020.
\newblock Abstractive summarization with combination of pre-trained
  sequence-to-sequence and saliency models.
\newblock \emph{arXiv preprint arXiv:2003.13028}.

\bibitem[{Serrano and Smith(2019)}]{DBLP:journals/corr/abs-1906-03731}
Sofia Serrano and Noah~A. Smith. 2019.
\newblock \href {http://arxiv.org/abs/1906.03731} {Is attention interpretable?}
\newblock \emph{CoRR}, abs/1906.03731.

\bibitem[{Shrikumar et~al.(2017)Shrikumar, Greenside, and
  Kundaje}]{pmlr-v70-shrikumar17a}
Avanti Shrikumar, Peyton Greenside, and Anshul Kundaje. 2017.
\newblock \href {https://proceedings.mlr.press/v70/shrikumar17a.html} {Learning
  important features through propagating activation differences}.
\newblock In \emph{Proceedings of the 34th International Conference on Machine
  Learning}, volume~70 of \emph{Proceedings of Machine Learning Research},
  pages 3145--3153. PMLR.

\bibitem[{Simonyan et~al.(2014)Simonyan, Vedaldi, and
  Zisserman}]{Simonyan14deepinside}
Karen Simonyan, Andrea Vedaldi, and Andrew Zisserman. 2014.
\newblock Deep inside convolutional networks: Visualising image classification
  models and saliency maps.
\newblock In \emph{In Workshop at International Conference on Learning
  Representations}.

\bibitem[{Song et~al.(2021)Song, Wang, Feng, and Liu}]{song2021new}
Kaiqiang Song, Bingqing Wang, Zhe Feng, and Fei Liu. 2021.
\newblock A new approach to overgenerating and scoring abstractive summaries.
\newblock \emph{arXiv preprint arXiv:2104.01726}.

\bibitem[{Sundararajan et~al.(2017)Sundararajan, Taly, and
  Yan}]{pmlr-v70-sundararajan17a}
Mukund Sundararajan, Ankur Taly, and Qiqi Yan. 2017.
\newblock \href {https://proceedings.mlr.press/v70/sundararajan17a.html}
  {Axiomatic attribution for deep networks}.
\newblock In \emph{Proceedings of the 34th International Conference on Machine
  Learning}, volume~70 of \emph{Proceedings of Machine Learning Research},
  pages 3319--3328. PMLR.

\bibitem[{Vaswani et~al.(2017)Vaswani, Shazeer, Parmar, Uszkoreit, Jones,
  Gomez, Kaiser, and Polosukhin}]{tfattention17Vaswani}
Ashish Vaswani, Noam Shazeer, Niki Parmar, Jakob Uszkoreit, Llion Jones,
  Aidan~N Gomez, \L~ukasz Kaiser, and Illia Polosukhin. 2017.
\newblock \href
  {http://papers.nips.cc/paper/7181-attention-is-all-you-need.pdf} {Attention
  is all you need}.
\newblock In I.~Guyon, U.~V. Luxburg, S.~Bengio, H.~Wallach, R.~Fergus,
  S.~Vishwanathan, and R.~Garnett, editors, \emph{Advances in Neural
  Information Processing Systems 30}, pages 5998--6008. Curran Associates, Inc.

\bibitem[{Wang et~al.(2018)Wang, Singh, Michael, Hill, Levy, and
  Bowman}]{wang-etal-2018-glue}
Alex Wang, Amanpreet Singh, Julian Michael, Felix Hill, Omer Levy, and Samuel
  Bowman. 2018.
\newblock \href {https://doi.org/10.18653/v1/W18-5446} {{GLUE}: A multi-task
  benchmark and analysis platform for natural language understanding}.
\newblock In \emph{Proceedings of the 2018 {EMNLP} Workshop {B}lackbox{NLP}:
  Analyzing and Interpreting Neural Networks for {NLP}}, pages 353--355,
  Brussels, Belgium. Association for Computational Linguistics.

\bibitem[{Wang et~al.(2016)Wang, Huang, Zhu, and
  Zhao}]{wang-etal-2016-attention}
Yequan Wang, Minlie Huang, Xiaoyan Zhu, and Li~Zhao. 2016.
\newblock \href {https://doi.org/10.18653/v1/D16-1058} {Attention-based {LSTM}
  for aspect-level sentiment classification}.
\newblock In \emph{Proceedings of the 2016 Conference on Empirical Methods in
  Natural Language Processing}, pages 606--615, Austin, Texas. Association for
  Computational Linguistics.

\bibitem[{Wiegreffe and Pinter(2019)}]{wiegreffe-pinter-2019-attention}
Sarah Wiegreffe and Yuval Pinter. 2019.
\newblock \href {https://doi.org/10.18653/v1/D19-1002} {Attention is not not
  explanation}.
\newblock In \emph{Proceedings of the 2019 Conference on Empirical Methods in
  Natural Language Processing and the 9th International Joint Conference on
  Natural Language Processing (EMNLP-IJCNLP)}, pages 11--20, Hong Kong, China.
  Association for Computational Linguistics.

\bibitem[{Wolf et~al.(2020)Wolf, Debut, Sanh, Chaumond, Delangue, Moi, Cistac,
  Rault, Louf, Funtowicz, Davison, Shleifer, von Platen, Ma, Jernite, Plu, Xu,
  Le~Scao, Gugger, Drame, Lhoest, and Rush}]{wolf-etal-2020-transformers}
Thomas Wolf, Lysandre Debut, Victor Sanh, Julien Chaumond, Clement Delangue,
  Anthony Moi, Pierric Cistac, Tim Rault, Remi Louf, Morgan Funtowicz, Joe
  Davison, Sam Shleifer, Patrick von Platen, Clara Ma, Yacine Jernite, Julien
  Plu, Canwen Xu, Teven Le~Scao, Sylvain Gugger, Mariama Drame, Quentin Lhoest,
  and Alexander Rush. 2020.
\newblock \href {https://doi.org/10.18653/v1/2020.emnlp-demos.6} {Transformers:
  State-of-the-art natural language processing}.
\newblock In \emph{Proceedings of the 2020 Conference on Empirical Methods in
  Natural Language Processing: System Demonstrations}, pages 38--45, Online.
  Association for Computational Linguistics.

\bibitem[{Xu and Durrett(2021)}]{xu-durrett-2021-dissecting}
Jiacheng Xu and Greg Durrett. 2021.
\newblock \href {https://doi.org/10.18653/v1/2021.acl-long.539} {Dissecting
  generation modes for abstractive summarization models via ablation and
  attribution}.
\newblock In \emph{Proceedings of the 59th Annual Meeting of the Association
  for Computational Linguistics and the 11th International Joint Conference on
  Natural Language Processing (Volume 1: Long Papers)}, pages 6925--6940,
  Online. Association for Computational Linguistics.

\bibitem[{Yang and Klein(2021)}]{yang-klein-2021-fudge}
Kevin Yang and Dan Klein. 2021.
\newblock \href {https://doi.org/10.18653/v1/2021.naacl-main.276} {{FUDGE}:
  Controlled text generation with future discriminators}.
\newblock In \emph{Proceedings of the 2021 Conference of the North American
  Chapter of the Association for Computational Linguistics: Human Language
  Technologies}, pages 3511--3535, Online. Association for Computational
  Linguistics.

\bibitem[{Zeiler and Fergus(2014)}]{zeiler2014visualizing}
Matthew~D. Zeiler and Rob Fergus. 2014.
\newblock Visualizing and understanding convolutional networks.
\newblock In \emph{Computer Vision -- ECCV 2014}, pages 818--833, Cham.
  Springer International Publishing.

\bibitem[{Zhang et~al.(2018)Zhang, Dinan, Urbanek, Szlam, Kiela, and
  Weston}]{zhang2018personalizing}
Saizheng Zhang, Emily Dinan, Jack Urbanek, Arthur Szlam, Douwe Kiela, and Jason
  Weston. 2018.
\newblock Personalizing dialogue agents: I have a dog, do you have pets too?
\newblock In \emph{Proceedings of the 56th Annual Meeting of the Association
  for Computational Linguistics (Volume 1: Long Papers)}, pages 2204--2213.

\bibitem[{Zhang et~al.(2019)Zhang, Kishore, Wu, Weinberger, and
  Artzi}]{zhang2019bertscore}
Tianyi Zhang, Varsha Kishore, Felix Wu, Kilian~Q Weinberger, and Yoav Artzi.
  2019.
\newblock Bertscore: Evaluating text generation with bert.
\newblock \emph{arXiv preprint arXiv:1904.09675}.

\bibitem[{Zhao et~al.(2017)Zhao, Zhao, and Eskenazi}]{zhao2017learning}
Tiancheng Zhao, Ran Zhao, and Maxine Eskenazi. 2017.
\newblock Learning discourse-level diversity for neural dialog models using
  conditional variational autoencoders.
\newblock \emph{arXiv preprint arXiv:1703.10960}.

\bibitem[{Zhao et~al.(2020)Zhao, Wu, Xu, Tao, Zhao, and
  Yan}]{zhao2020knowledge}
Xueliang Zhao, Wei Wu, Can Xu, Chongyang Tao, Dongyan Zhao, and Rui Yan. 2020.
\newblock Knowledge-grounded dialogue generation with pre-trained language
  models.
\newblock \emph{arXiv preprint arXiv:2010.08824}.

\bibitem[{Zhong et~al.(2021)Zhong, Friedman, and
  Chen}]{DBLP:journals/corr/abs-2104-05240}
Zexuan Zhong, Dan Friedman, and Danqi Chen. 2021.
\newblock \href {http://arxiv.org/abs/2104.05240} {Factual probing is {[MASK]:}
  learning vs. learning to recall}.
\newblock \emph{CoRR}, abs/2104.05240.

\end{thebibliography}
